\documentclass[10pt,twocolumn,letterpaper]{article}

\usepackage{cvpr}      

\usepackage{graphicx}
\usepackage{amsmath}
\usepackage{amssymb}
\usepackage{booktabs}
\usepackage[noend]{algpseudocode}

\usepackage{algorithmicx,algorithm}

\usepackage[pagebackref,breaklinks,colorlinks]{hyperref}

\usepackage[capitalize]{cleveref}
\crefname{section}{Sec.}{Secs.}
\Crefname{section}{Section}{Sections}
\Crefname{table}{Table}{Tables}
\crefname{table}{Tab.}{Tabs.}

\usepackage{color}
\definecolor{runpei-blue}{RGB}{0, 113, 188}

\begin{document}

\title{MatrixVT: Efficient Multi-Camera to BEV Transformation for 3D Perception}


\author{
    Hongyu Zhou,~
    Zheng Ge,~
    Zeming Li,~
    Xiangyu Zhang\\
    MEGVII Technology\\
    {\tt\small\{zhouhongyu,gezheng,lizeming,zhangxiangyu\}@megvii.com}
}

\maketitle
\begin{abstract}

This paper proposes an efficient multi-camera to Bird's-Eye-View (BEV) view transformation method for 3D perception, dubbed MatrixVT.
Existing view transformers either suffer from poor transformation efficiency or rely on device-specific operators, hindering the broad application of BEV models. In contrast, our method generates BEV features efficiently with only convolutions and matrix multiplications (MatMul). Specifically, we propose describing the BEV feature as the MatMul of image feature and a sparse Feature Transporting Matrix (FTM). A Prime Extraction module is then introduced to compress the dimension of image features and reduce FTM's sparsity. Moreover, we propose the Ring \& Ray Decomposition to replace the FTM with two matrices and reformulate our pipeline to reduce calculation further. Compared to existing methods, MatrixVT enjoys a faster speed and less memory footprint while remaining deploy-friendly. Extensive experiments on the nuScenes benchmark demonstrate that our method is highly efficient but obtains results on par with the SOTA method in object detection and map segmentation tasks.

\end{abstract}

\section{Introduction}
\label{sec:intro}

\begin{figure}
    \centering
    \includegraphics[width=\columnwidth]{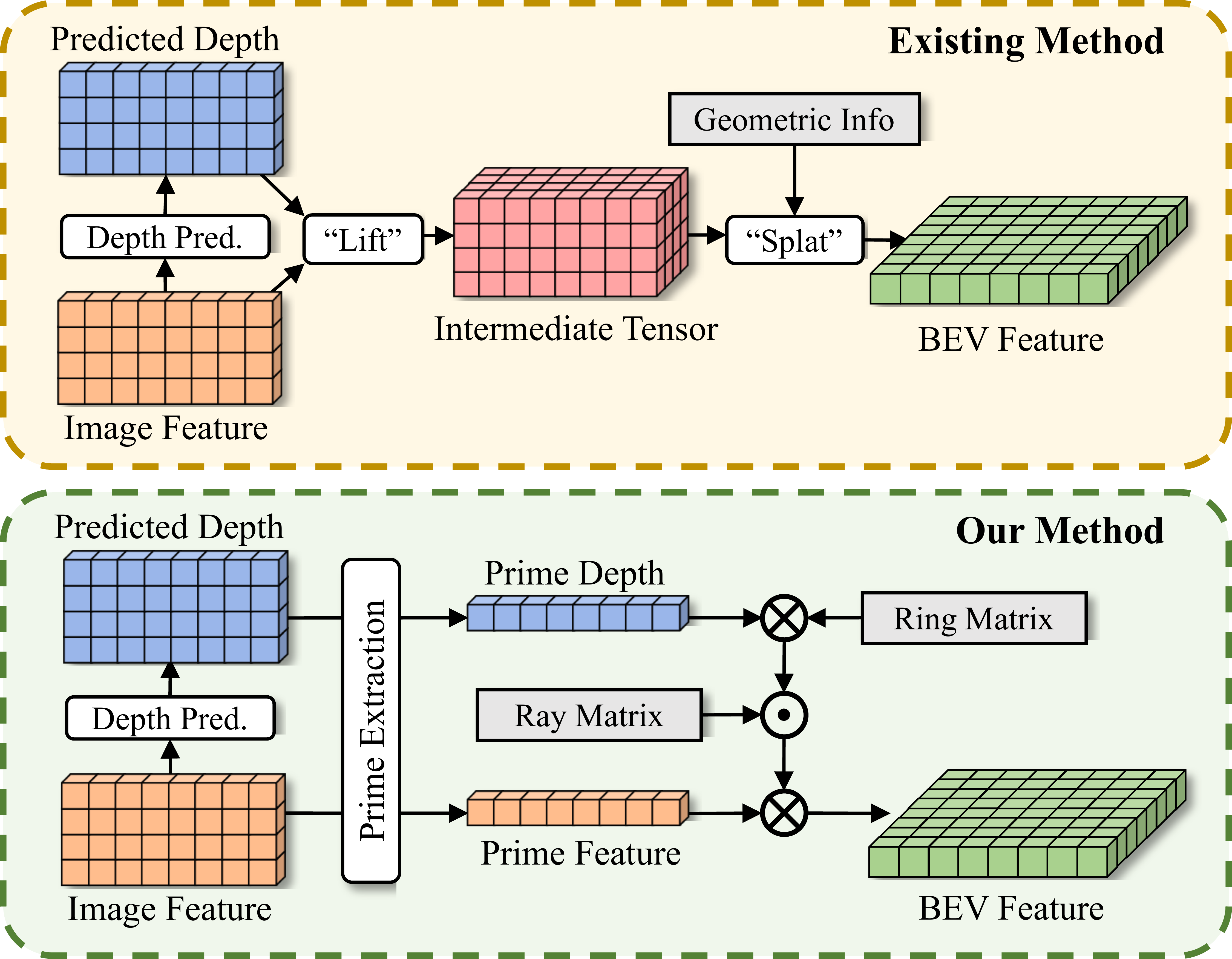}
    \caption{The pipeline of Lift-Splat (upper) and our proposed MatrixVT (lower). We compress image features before VT to reduce memory footprint and calculation. Note that only standard operators are adopted in MatrixVT.}
    \label{fig:compare}
\end{figure}

Vision-centric 3D perception in Bird's-Eye-View (BEV)~\cite{lss,pon,birdgan} has recently drawn extensive attention. Apart from their outstanding performance, the compact and unified feature representation in BEV facilitates straight-forward feature fusions~\cite{bevdet4d,bevfusion,bevformer,sts}, and enables various downstream tasks (\eg object detection~\cite{bevdet,bevdepth,bevformer}, map segmentation~\cite{lss,m2bev}, motion planning, \etc.) to be applied thereon easily. 

View Transformation (VT) is the key component that converts multi-camera features to BEV, which has been heavily studied in previous works~\cite{vpn,pon,lss,birdgan,bevformer,bevdet,oft,bevdepth}.
Existing VT methods can be categorized into \emph{geometry-based}~\cite{ipm,cam2bev,lss,bevdepth,caddn,m2bev} and \emph{learning-based} methods~\cite{bevformer,pon,vpn}. Among these two categories, geometry-based methods show superior performance due to the use of geometric constraints. \emph{Lift-Splat}~\cite{lss}, as a representative geometry-based VT, predicts \emph{categorical depth distribution} for each pixel and ``lift'' the corresponding features into 3D space according to the predicted depth. These feature vectors are then \emph{gathered} into pre-defined grids on a reference BEV plane (\ie, ``splat'') to form the BEV feature (Fig.~\ref{fig:compare}, upper). The Lift-Splat-based VT has shown great potential to produce high-quality BEV features, achieving remarkable performance on object detection~\cite{bevdepth, bevdet4d} and map segmentation tasks on the nuScenes benchmark~\cite{nuscenes}.

Despite the effectiveness of Lift-Splat-like VT~\cite{bevdepth}, two issues remain. First, the ``splat'' operation is not universally feasible. 
Existing implementations of ``splat'' relies on either the ``cumsum trick''~\cite{lss,bevdet} that is highly inefficient, or customized operators~\cite{bevdepth} that can only be used on specific devices, increases the cost of applying BEV perception. Second, the size of ``lifted'' multi-view image features is \emph{huge}, becoming the memory bottleneck of BEV models. These two issues lead to a heavy burden on BEV methods during both the training and inference phases. 
\emph{As a result, the drawbacks of existing view transformers limit the broad application of autonomous driving technology}

In this work, we propose a novel VT method, MatrixVT, to address the above problems. MatrixVT is proposed based on the fact that the VT can be viewed as a \emph{feature transportation} process. In that case, the BEV feature can be viewed as the MatMul between the ``lifted'' feature and a transporting matrix, namely \emph{Feature Transporting Matrix (FTM)}. We thus generalize the Lift-Splat VT into a purely mathematical form and eliminate specialized operators. 

However, transformation with FTM is a kind of degradation --- the mapping between the 3D space and BEV grids is \emph{extreme sparse}, leading to the huge size of FTM and poor efficiency. Prior works~\cite{bevdepth,caddn} seek customized operators, successfully avoiding such sparsity. In this paper, we argue that there are other solutions to the problem of sparse mapping.
First, we propose Prime Extraction. Motivated by an observation that the height (vertical) dimension of images is \emph{less} informative in autonomous driving (see Sec.~\ref{sec:compress}), we compress the image features along this dimension before VT. 
Second, we adopt matrix decomposition to reduce the sparsity of FTM. The proposed Ring \& Ray Decomposition orthogonally decomposes the FTM into two separate matrices, each encoding the distance and direction of the ego-centric polar coordinate. This decomposition also allows us to reformulate our pipeline into a \emph{mathematically equivalent} but more efficient one (Fig.~\ref{fig:compare}, lower). These two techniques reduce memory footprint and calculation during VT by hundreds of times, enabling MatrixVT to be more efficient than existing methods.

The proposed MatrixVT inherits the advantages of the Lift-Splat~\cite{lss} paradigm while being much more memory efficient and fast.
Extensive experimental results show that MatrixVT is 2-to-8 times faster than previous methods~\cite{bevdepth,bevdet} and saves up to 97\% memory footprint among different settings. Meanwhile, the perception model with MatrixVT achieves 46.6\% mAP and 56.2\% NDS for object detection and 46.2\% mIoU for vehicle segmentation on the nuScenes~\cite{nuscenes} val set, which is comparable to the state-of-the-art performance~\cite{bevdepth}. 
We conclude our main contributions as follows:
\begin{itemize}
    \item We propose a new description of multi-camera to BEV transformation --- using the Feature Transportation Matrix (FTM), which is a more general representation.
    \item To solve the sparse mapping problem, we propose Prime Extraction and the Ring \& Ray Decomposition, boosting VT with FTM by a huge margin.
    \item Extensive experiments
    demonstrate that MatrixVT yields comparable performance to the state-of-the-art method on the nuScenes object detection and map segmentation tasks while being more efficient and generally applicable.

\end{itemize}


\section{Related Works}
\subsection{Visual 3D Perception}
Perception of 3D objects and scenes takes a key role in autonomous driving and robotics, thus attracting increasing attention nowadays. Camera-based perception~\cite{fcos3d,pgd,bevdet,bevformer,petr} is the most commonly used method for varies scenarios due to its low cost and high accessibility. Comparing with 2D perception (object detection~\cite{fcos}, semantic segmentation~\cite{unet}, \textit{etc.}), 3D perception requires additional prediction of the depth information which is an naturally ill-posed problem~\cite{oft}. 

Existing works either predict the depth information explicitly or implicitly. FCOS3D~\cite{fcos3d} simply extend the structure of the classic 2D object detector~\cite{fcos}, predicting pixel-wise depth explicitly using an extra sub-net that is supervised by LiDAR data. CaDDN~\cite{caddn} propose treating depth prediction as classification task rather than regression task, and project image feature into Bird's-Eye-View (BEV) space to achieve unified modeling of detection and depth prediction. BEVDepth and following works~\cite{bevdepth,bevstereo,sts} propose several techniques to enhance depth prediction, these works achieve outstanding performance due to precise depth. Meanwhile, methods like PON~\cite{pon} and BirdGAN~\cite{birdgan}  use pure neural networks to transform image features into BEV space,  learning object depth implicitly using segmentation or detection supervision.  
Currently, methods that explicitly learn depth show prior performance than implicit approaches thanks to the supervision from LiDAR data and depth modeling. In this paper, we use the DepthNet same as in BEVDepth for high-performance.

\subsection{Perception in Bird's-Eye-View}
The concept of BEV is firstly proposed for processing LiDAR point cloud~\cite{voxelnet,pointpillars,second,centerpoint}, and found effective for fusing multi-view image features. The core component of vision-based BEV paradigm is the view-transformation. OFT~\cite{oft} firstly propose mapping image feature from Perspective View into BEV using camera parameters. This method project a reference point from an BEV grid to image plane, and sample corresponding features back to the BEV grid. Following this work, BEVFormer~\cite{bevformer} propose using Deformable Cross Attention to sample features around the reference point. These methods fail to distinguish BEV grids that are projected to same position on the image plane, thus show inferior performance than depth-based methods.

Depth-based methods, represented by LSS~\cite{lss} and CaDDN~\cite{caddn}, predict categorical depth for each pixel, the extracted image feature on a specific pixel is then projected into 3D space by doing per-pixel outer product with corresponding depth. The projected high-dimensional tensor is then ``collapsed'' to a BEV reference plane using convolution~\cite{caddn}, Pillar Pooling~\cite{lss,bevdet}, or Voxel Pooling~\cite{bevdepth}. Lift-Splat based methods~\cite{lss,bevdepth,bevdet,bevstereo} show outstanding performance for down-stream tasks, but introduces two extra problems. Firstly, the intermediate representation of image feature is large and in-efficient, making training and application of these methods difficult. Secondly, the Pillar Pooling~\cite{lss} introduces random memory access, which is slow and device demanding (extremely slow on general-purpose devices). In this paper, we propose a new depth-based view transformation to overcome these problems while retaining the ability of producing high-quality BEV features.
\begin{figure}
    \centering
    \includegraphics[width=0.95\columnwidth]{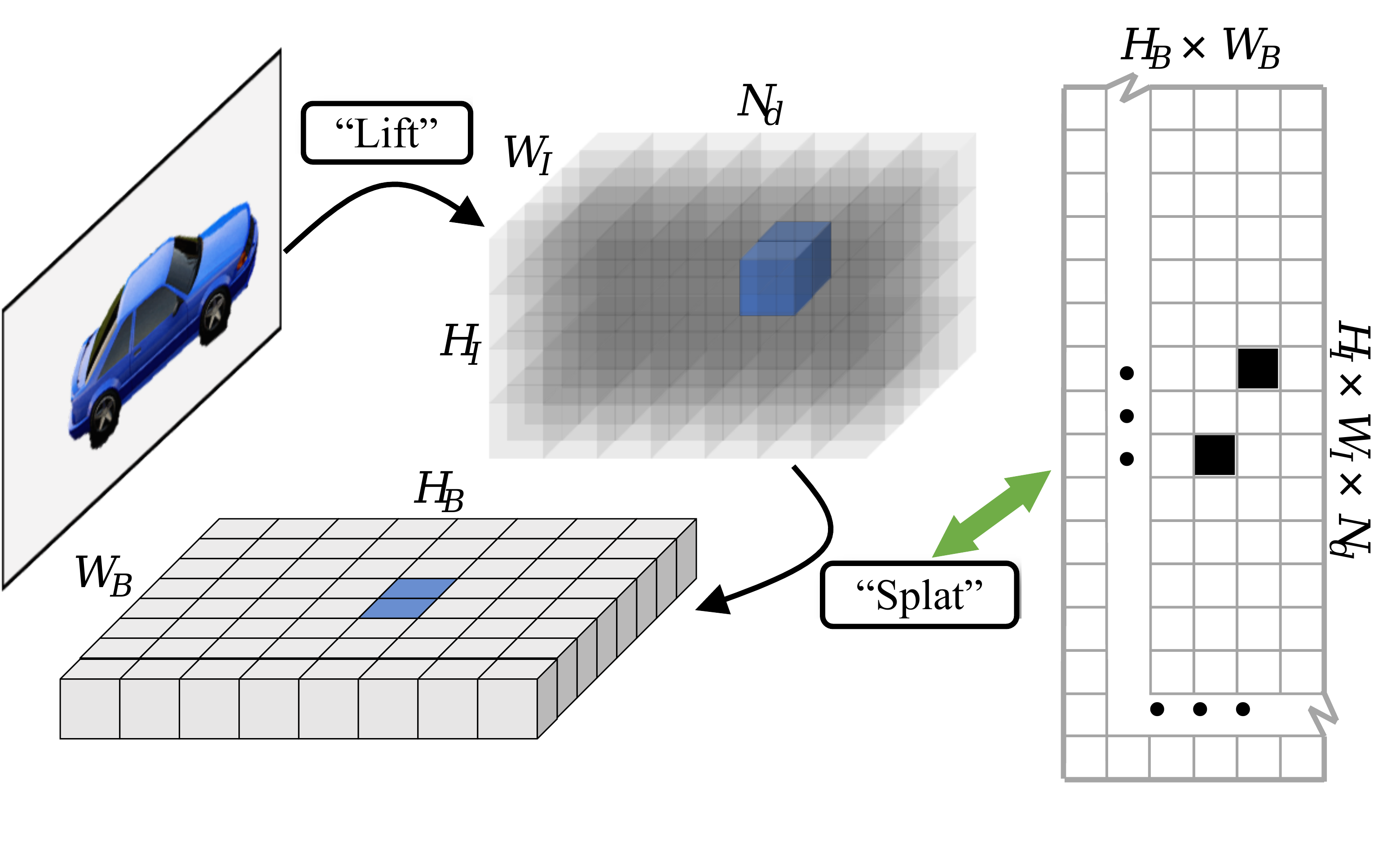}    
    \caption{The View Transformation can be implemented by standard operators since the ``splat'' operation in Lift-Splat-like transformations can be represented by a binary matrix.}
    \label{fig:trans}
\end{figure}

\section{MatrixVT}
Our MatrixVT is a simple view transformer based on the depth-based VT paradigm.
In Sec.~\ref{sec:FTM}, we first revisit existing Lift-Splat transformation~\cite{lss,bevdepth} and introduce the concept of Feature Transporting Matrix (FTM) together with the sparse mapping problem.
Then, techniques proposed to solve the problem of sparse mapping \ie, Prime Extraction (Sec.~\ref{sec:compress}) and Ring \& Ray Decomposition (Sec.~\ref{sec:rr}), are introduced. In Sec.~\ref{sec:ours}. we designate the novel VT method utilizing the aforementioned techniques as MatrixVT, and elaborate its overall pipeline. 

For clarity, we use letters of the normal script (\eg $F$) for tensors, the Roman script for matrices (\eg $\mathrm{M}$), and Bold script (\eg $\boldsymbol{v}$) for vectors. Besides, $\mathrm{A} \cdot \mathrm{B}$ denotes Matrix Multiplication (MatMul), $\mathrm{A} \odot \mathrm{B}$ denotes Hadamard Product, and $\boldsymbol{a} \otimes \boldsymbol{b}$ denotes outer product.
\subsection{Background}
\label{sec:FTM}

Lift-Splat-like methods exploit the idea of pseudo-LiDAR~\cite{pseudolidar} and LiDAR-based BEV methods~\cite{voxelnet,pointpillars} for View Transformation. Image features are ``lifted'' to 3D space and being processed like pointcloud.
We define $N_c$ as the number of cameras for a specific scene; $N_d$ as the number of depth bins; $H_I$ and $W_I$ to be the height and width of image features; $H_B$ and $W_B$ be the height and width of BEV features (\ie, shape of BEV grids); $C$ to be the number of feature channels. Consequently, let multi-view image features to be $F \in \mathbb{R}^{N_c \times H_I \times W_I \times C}$; categorical depth to be $D \in \mathbb{R}^{N_c \times H_I \times W_I \times D}$; BEV features to be $F_{BEV} \in \mathbb{R}^{H_B \times W_B \times C}$ (initialized to zeros). 
The Lift-Splat VT first ``lift'' the $F$ into 3D space by doing per-pixel outer product with $D$, obtaining the high-dimensional intermediate tensor $F_{inter} \in \mathbb{R}^{N_c \times H_I \times W_I \times N_d \times C}$.
\begin{equation}
    F_{inter}=\{F^{h,w}_{inter}\}_{H_I,W_I}=\{\boldsymbol{F^{h,w}} \cdot (\boldsymbol{D^{h,w}})^T\}_{H_I,W_I}
\end{equation}
The intermediate tensor can be treated as $N_c \times H_I \times W_I \times N_d$ feature vectors, each vector corresponding to a geometric coordinate. The ``splat'' operation is then adopted using operators like Pillar Pooling~\cite{lss}, during which each feature vector is summed to a BEV grid according to geometric coordinates (see Fig.~\ref{fig:trans}). 

We find that the ``splat'' operation is a fixed mapping between the intermediate tensor and BEV grids. Therefore, we use a Feature Transporting Matrix $\mathrm{M_{FT}} \in \mathbb{R}^{(H_B \times W_B) \times (N_c \times H_I \times W_I \times N_d)}$ to describe this mapping. The FTM can be strictly equal to Pillar Pooling in LSS~\cite{lss} or encodes different sampling rules (\ie, Gaussian sampling). The FTM enables feature transportation with Matrix Multiplication (MatMul):
\begin{equation}
    \mathrm{F_{BEV}} = \mathrm{M_{FT}} \cdot \mathrm{F_{inter}}
\end{equation}
where $\mathrm{F_{inter}} \in \mathbb{R}^{(N_c \times H_I \times W_I \times N_d) \times C}$ is the ``lifted'' feature $F_{inter}$ in matrix format, $\mathrm{F_{BEV}} \in \mathbb{R}^{(H_B \times W_B) \times C}$ is the BEV feature $F_{BEV}$ in matrix format.

Replacing the ``splat'' operation with FTM eliminates the need for customized operators. However, the sparse mapping between the 3D space and BEV grids can lead to a \emph{massive} and \emph{highly sparse} FTM, harm the efficiency of matrix-based VT. To address the sparse mapping problem without customized operators, we propose two techniques to reduce the sparsity of FTM and speed up the transformation.

\subsection{Prime Extraction for Autonomous Driving}
\label{sec:compress}
\begin{figure}
    \centering
    \includegraphics[width=\columnwidth]{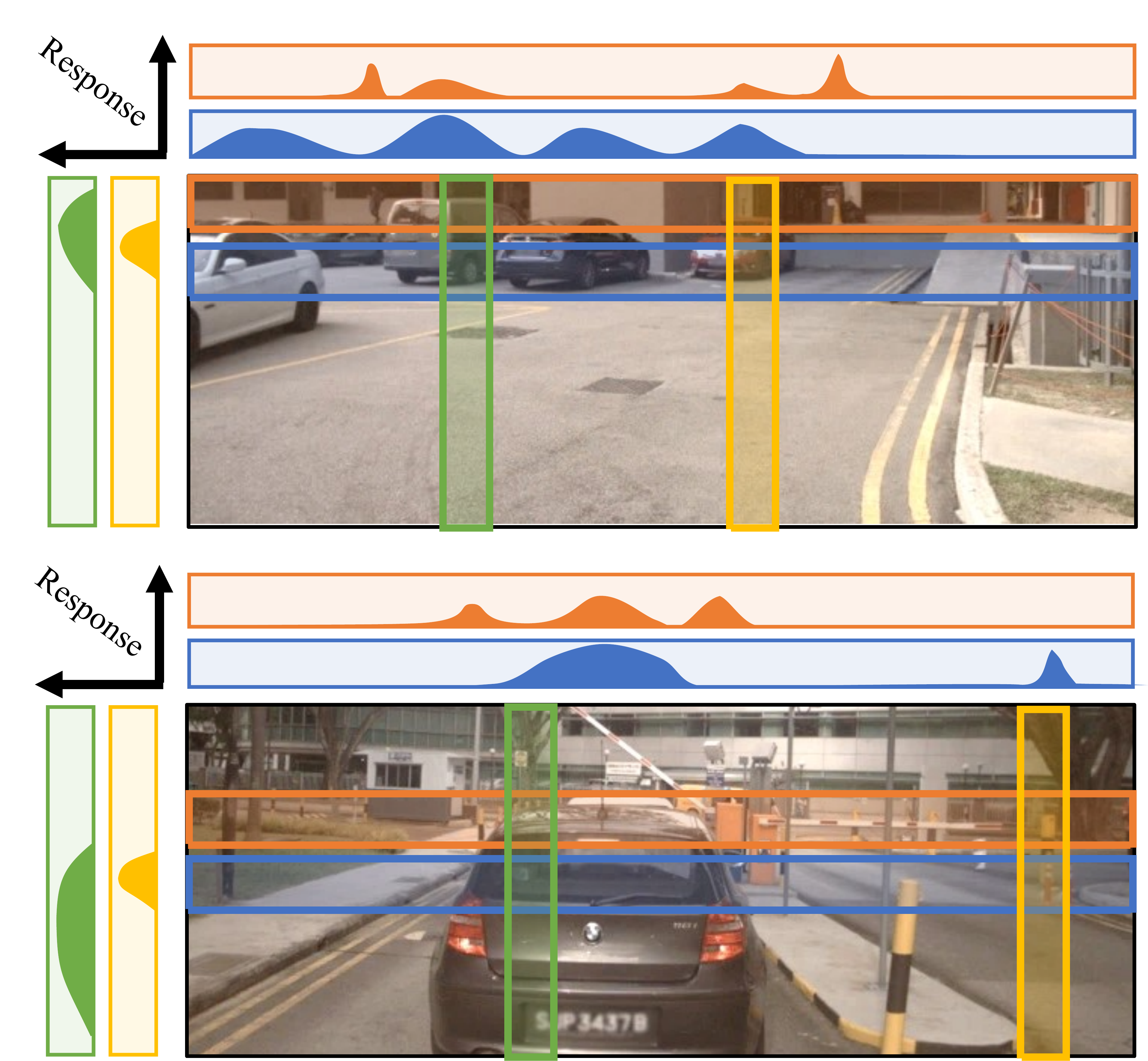}
    \caption{Response strengths along vertical and horizontal dimensions. For autonomous driving, the variance of response on the width dimension is higher than height dimension.}
    \label{fig:redun}
\end{figure}
The high-dimensional intermediate tensor is the primary cause of the sparse mapping and the low efficiency of Lift-Splat-like VT. An intuitive way to reduce sparsity is reducing the size of $F_{inter}$. Therefore, we propose Prime Extraction --- a compression technique for autonomous driving and other scenarios where information redundancy exists. 

\begin{figure}
    \centering
    \includegraphics[width=0.95\columnwidth]{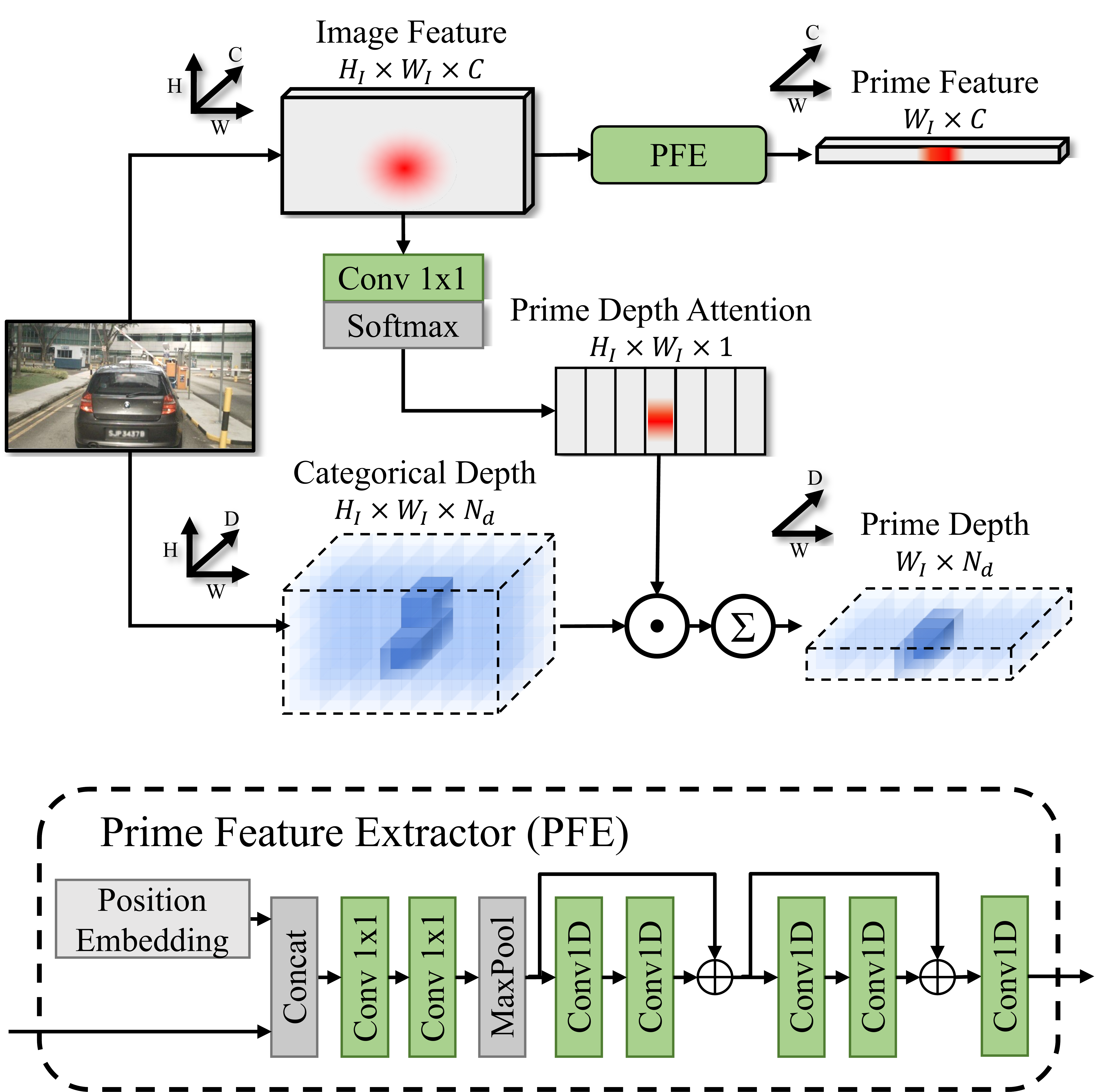}
    \caption{The Prime Extraction module. The categorical depth is reduced guided by Prime Depth Attention, while the image feature is reduced by Prime Feature Extractor (PFE), which consists of MaxPooling and convolutions for refinement.}
    \label{fig:pe}
\end{figure}

\begin{figure*}[t]
    \centering
    \includegraphics[width=\textwidth]{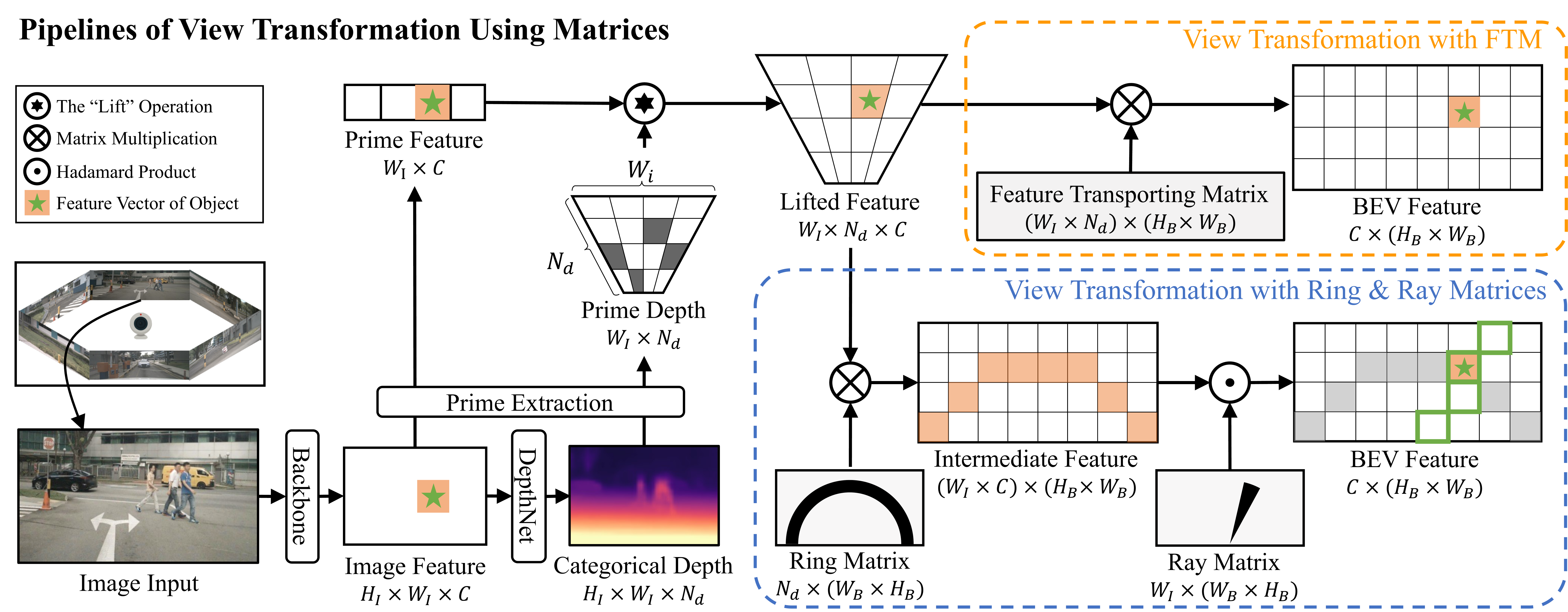}
    \caption{View Transformation using matrices. We first extract features from the image and predict categorical depth for each pixel. The obtained feature and depth are sent to the Prime Extraction module for compression. With the compressed Prime Feature and the Prime Depth, we ``lift'' the 1D Prime Feature into 2D space using Prime Depth. The ``lifted feature'' is then transformed into the BEV feature by MatMul with FTM or the decomposed Ring \& Ray matrices.}
    \label{fig:pipeline}
\end{figure*}

As shown in Fig.~\ref{fig:pe}, the Prime Extraction module predicts Prime Depth Attention for each direction (a column of the tensor) guided by the image feature. It generates the Prime Depth as a \emph{weighted sum} of categorical depth distributions. Meanwhile, the Prime Feature is obtained by the Prime Feature Extractor (PFE), which consists of position embedding, column-wise max-pooling, and convolutions for refinement. By applying Prime Extraction to matrix-based VT, we successfully reduce the FTM to $\mathrm{M'_{FT}} \in \mathbb{R}^{(H_B \times W_B) \times (N_c \times W_I \times N_d)}$, which is $H_I$ times smaller than the raw matrix. The pipeline of generating the BEV feature using the Prime Feature and the Prime Depth is demonstrated in Fig.~\ref{fig:pipeline} (yellow box).


The prime extraction is motivated by an observation: The image feature's height dimension has a lower response variance than the width dimension. This observation indicates that this dimension contains less information than the width dimension. We thus propose to compress image features on the height dimension. Previous works have also exploited reducing the height dimension of image features~\cite{pon,transfusion}, but we firstly propose compressing both image features and corresponding depth to boost VT.

In Sec.~\ref{sec:pe-exp}, we will show that the extracted Prime Feature and Prime Depth effectively retain valuable information from the raw feature and produce BEV features of the same high quality as the raw feature. Moreover, the Prime Extraction technique can be individually adopted to existing Lift-Splat-like VTs to enhance their efficiency at almost no performance cost.

\subsection{``Ring and Ray'' Decomposition}
\label{sec:rr}
The sparsity FTM can be further reduced by matrix decomposition. To this end, we propose the ``Ring \& Ray'' Decomposition. Without loss of generality, we can set the $N_c$ to 1 and view $\mathrm{M'_{FT}}$ as a tensor $M'_{FT}$. In that case, the shape of $M'_{FT}$ would be $H_B \times W_B \times W_I \times N_d$. We note that its dimension of size $W_I$ can be viewed as the \emph{direction} in a \emph{polar coordinate}, since each column of the image feature represents information of a specific direction. Likewise, the dimension of size $N_d$ can be viewed as the \emph{distance} in the polar coordinate. In other words, the image feature required for a specific BEV grid can be located by direction and distance. We thus propose to orthogonally decompose the $M'_{FT}$ into two separate matrices, each encoding the directional or distance information. Specifically, we use a Ring Matrix $\mathrm{M_{Ring}} \in \mathbb{R}^{N_d \times (H_B \times W_B)}$ to encode distance information and a Ray Matrix $\mathrm{M_{Ray}} \in \mathbb{R}^{W_I \times (H_B \times W_B)}$ to encode directional information (see \emph{Appendix 1.1} for pseudo code). The Ring \& Ray Decomposition effectively reduces the size of static parameters. The number of predefined parameters (size of FTM) is reduced from $W_I \times N_d \times H_B \times W_B$ to $(W_I + N_d) \times H_B \times W_B$, which is typically 30 to 50 times less\footnote{Under feature width 44 and 88, with 112 depth bins.}.

Then we show how to use these two matrices for VT. Given the Prime Feature and Prime Depth, we first do the per-pixel outer product of them as in Lift-Splat to obtain the ``lifted feature'' $F_{inter} \in \mathbb{R}^{W_I \times N_d \times C}$. 
\begin{equation}
\label{eq:lift}
    F_{inter}=\{F^w_{inter}\}_{W_I}=\{\boldsymbol{F^w} \otimes (\boldsymbol{D^w)^T}\}_{W_I}
\end{equation}
Then, as illustrated in Fig.~\ref{fig:pipeline}, instead of using $\mathrm{M'_{FT}}$ to directly transform it into a BEV feature (yellow box), we transpose $F_{inter}$, view it as a matrix $\mathrm{F_{inter}} \in \mathbb{R}^{N_d \times (W_I \times C)}$, and do MatMul between the Ring Matrix and $\mathrm{F_{inter}}$ to obtain an intermediate feature. The intermediate feature is then masked by doing a Hadamard Product with the Ray Matrix, summed on the dimension of $W_I$ to obtain the BEV feature. (Fig.~\ref{fig:pipeline}, blue box, summation omitted).
\begin{gather}
    \mathrm{F^*_{inter}}=\mathrm{M_{Ray}} \odot (\mathrm{M_{Ring}} \cdot \mathrm{F_{inter}})\\
    F_{BEV} = \sum_w F^{*w}_{inter}
\label{eq:proc}
\end{gather}
where $F^*_{inter} \in \mathbb{R}^{W_I \times C \times H_B \times W_B}$ is $\mathrm{F^*_{inter}}$ in tensor form.

However, this decomposition do not reduce the FLOPs during VT and introduces the Intermediate Feature in Fig.~\ref{fig:pipeline} (blue) whose size if huge and depends on feature channel $C$. To reduce the calculation and memory footprint during VT, we combine Eq.~\ref{eq:lift} to Eq.~\ref{eq:proc} and rewrite them in a mathematically equivalent form (see \emph{Appendix 1.2} for proof):
\begin{equation}
\label{eq:matrixvt}
    \mathrm{F_{BEV}}=(\mathrm{M_{Ray}} \odot (\mathrm{M_{Ring}} \cdot \mathrm{D})) \cdot \mathrm{F_I}
\end{equation}
With this reformulation, we reduce the calculation during VT from $2 \times W_I \times C \times N_d \times H_B \times W_B \text{~FLOPs}$ to $2(C + N_d + 1) \times W_I \times H_B \times W_B \text{~FLOPs}$; the memory footprint is also reduced from $W_I \times N_d \times H_B \times W_B$ to $(W_I + N_d) \times H_B \times W_B$. Under common setting ($C=80, N_d=112, W_I=44$), the Ring \& Ray Decomposition reduces calculation by 46 times and saves 96\% memory footprint.

\begin{table*}
	\begin{center}
	\scalebox{0.95}{	
  \begin{tabular}{l|c|c|c|c|ccccc|c}
\toprule
    Method  & Backbone & Resolution & MF & mAP$\uparrow $ & mATE$\downarrow$ & mASE$\downarrow$ & mAOE$\downarrow$ & mAVE$\downarrow$ & mAAE$\downarrow$ & NDS$\uparrow $\\
\midrule
BEVDet~\cite{bevdet}
    & Res-50 & $256\times704$ & $\times$ & 0.286 & 0.724 & 0.278 & 0.590 & 0.873 & 0.247 & 0.372\\
PETR~\cite{petr}
    & Res-50 & $384\times1056$ & $\times$ & 0.313 & 0.768 & 0.278 & 0.564 & 0.923 & 0.225 & 0.381\\
BEVDepth~\cite{bevdepth}
    & Res-50 & $256\times704$ & $\times$ & \textbf{0.337} & 0.646 & 0.271 & 0.574 & 0.838 & 0.220 & 0.414\\
MatrixVT
    & Res-50 & $256\times704$ & $\times$ & 0.336 & 0.653 & 0.271 & 0.473 & 0.903 & 0.231 & \textbf{0.415}\\
\midrule
BEVDet~\cite{bevdet}
    & Res-101 & $384\times1056$ & $\times$ & 0.330 & 0.702 & 0.272 & 0.534 & 0.932 & 0.251 & 0.396\\
FCOS3D~\cite{fcos3d}
    & Res-101 & $900\times1600$ & $\times$ & 0.343 & 0.725 & 0.263 & 0.422 & 1.292 & 0.153 & 0.415\\
PETR~\cite{petr}
    & Res-101 & $512\times1408$ & $\times$ & 0.357 & 0.710 & 0.270 & 0.490 & 0.885 & 0.224 & 0.421\\
BEVFormer~\cite{bevformer}
    & R101-DCN & $900\times1600$ & $\times$ & 0.375 & 0.725 & 0.272 & 0.391 & 0.802 & 0.200 & 0.448\\
MatrixVT
    & Res-101 & $512\times1408$ & $\times$ & \textbf{0.396} & 0.577 & 0.261 & 0.397 & 0.870 & 0.207 & \textbf{0.467}\\
\midrule
BEVFormer~\cite{bevformer}
    & R101-DCN & $900\times1600$ & $\checkmark$ & 0.416 & 0.673 & 0.274 & 0.372 & 0.394 & 0.198 & 0.517\\
BEVDet4D~\cite{bevdet4d}
    & Swin-B & $640\times1600$ & $\checkmark$ & 0.421 & 0.579 & 0.258 & 0.329 & 0.301 & 0.191 & 0.545\\
BEVDepth~\cite{bevdepth}
    & V2-99 & $512\times1408$ & $\checkmark$ & 0.464 & 0.528 & 0.254 & 0.350 & 0.354 & 0.198 & \textbf{0.564}\\
MatrixVT
    & V2-99 & $512\times1408$ & $\checkmark$ & \textbf{0.466} & 0.535 & 0.260 & 0.380 & 0.342 & 0.198 & 0.562\\
\bottomrule
  \end{tabular}
  }
\caption{Experimental results of object detection on the nuScenes val set. The ``MF'' indicates multi-frame fusion.}
\label{table:det}
	\end{center}
\end{table*}

\subsection{Overall Pipeline}
\label{sec:ours}
With above techniques, we reduce the calculation and memory footprint using FTM by hundreds times, making VT with FTM not only feasible but also efficient. 
Given the multi-view images for a specific scene, we conclude the overall pipeline of MatrixVT as follows: 
\begin{enumerate}
    \item We first use an image backbone to extract image features from each image.
    \item Then, a depth predictor is adopted to predict categorical depth distribution for each feature pixel to obtain the depth prediction.
    \item After that, we send each image feature and corresponding depth to the Prime Extraction module, obtaining the Prime Feature and the Prime Depth, which is the compressed feature and depth.
    \item Finally, with the Prime Feature, Prime Depth, and the pre-defined Ring \& Ray Matrices, we use Eq.~\ref{eq:matrixvt} (see also Fig.~\ref{fig:compare}, lower) to obtain the BEV feature.
\end{enumerate}

\section{Experiments}
In this section, we compare the performance, latency, and memory footprint of MatrixVT and other existing VT methods on the nuScenes benchmark~\cite{nuscenes}.
\subsection{Implementation Details}
We conduct our experiments based on BEVDepth~\cite{bevdepth}, the current state-of-the-art detector on the nuScenes benchmark. In order to conduct a fair comparison of performance and efficiency, we re-implement BEVDepth according to their paper.
Unless otherwise specified, we use ResNet-50~\cite{resnet} and VoVNet2-99~\cite{vovnet2} pre-trained on DD3D~\cite{dd3d} as the image backbone and SECOND FPN~\cite{second} as the image neck and BEV neck. 
The input image adopts pre-processing and data augmentations same as in \cite{bevdepth}. We use BEV feature size $128 \times 128$ for low input resolution and $256 \times 256$ for high resolution on detection. Segmentation experiments use $200 \times 200$ BEV resolution as in LSS~\cite{lss}. We use the DepthNet~\cite{bevdepth} to predict categorical depth from 2m to 58m in nuScenes, with uniform 112 division. During training, CBGS~\cite{cbgs} and model EMA are adopted. Models are trained to converge since MatrixVT converges a little slower than other methods, but no more than 30 epochs.

\subsection{Comparison of Performances}
\subsubsection{Object Detection}
We conduct experiments under several settings to evaluate the performance of MatrixVT in Tab.~\ref{table:det}. We first adopt the ResNet family as the backbone without applying multi-frame fusion. MatrixVT achieves 33.6\% and 49.7\% mAP with ResNet-50 and ResNet-101, which is comparable to the BEVDepth~\cite{bevdepth} and surpasses other methods by a large margin. We then test the upper bound of MatrixVT by replacing the backbone with V2-99~\cite{vovnet2} pre-trained on external data and applying multi-frame fusion. 
Under this setting, MatrixVT achieves 46.6\% mAP and 56.2\% NDS, which is also comparable to the BEVDepth. 

\begin{figure*}
    \centering
    \includegraphics[width=\textwidth]{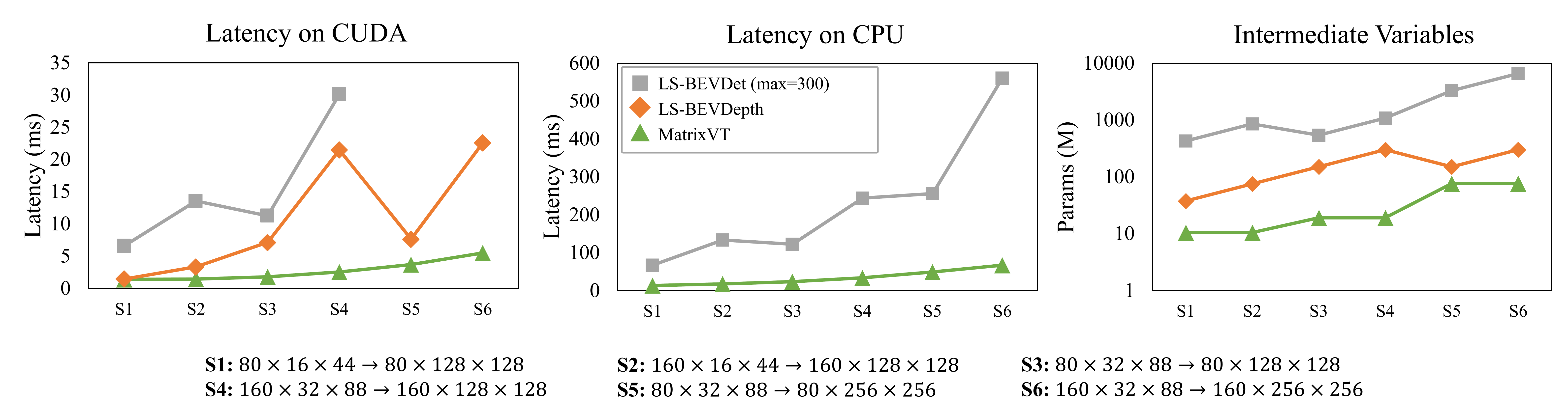}
    \caption{Latency (left, middle) and the number of intermediate parameters (right) of MatrixVT and two implementations of Lift-Splat, the S1$\sim$S6 denote transformation settings represented by image feature size ($C \times H_I \times W_I$) to BEV feature size ($C \times H_B \times W_B$). Note that the LS-BEVDet raises Out Of Memory error under S5 and S6 on CUDA. The LS-BEVDepth is not available without the CUDA platform.}
    \label{fig:latency}
\end{figure*}

\subsubsection{Map Segmentation}
We also conduct experiments on map segmentation tasks to validate the quality of the BEV feature generated by MatrixVT. To achieve this, we simply put a U-Net-like~\cite{unet} segmentation head on the BEV feature. For fair comparison, we put the same head on the BEV feature of BEVDepth for experiments, and results are reported as ``BEVDepth-Seg'' in Tab.~\ref{table:seg}. It is worth noting that previous works achieve the best segmentation performance under different settings (different resolution, head structure, \etc); thus, we report the highest performance of each method. 
As can be seen from Tab.~\ref{table:seg}, the map segmentation performance of MatrixVT surpasses most existing methods on all three sub-tasks and is comparable to our baseline, BEVDepth.
\begin{table}
	\begin{center}
		\scalebox{0.9}{
  \begin{tabular}{c|ccc}
\toprule
    Method  & IoU-Drive$\uparrow$ & IoU-Lane$\uparrow$ & IoU-Vehicle$\uparrow$ \\
\midrule
LSS~\cite{lss}
    & 0.729 & 0.200 & 0.321\\
FIERY~\cite{fiery}
    & - & - & 0.382\\
M\textsuperscript{2}BEV~\cite{m2bev}
    & 0.759 & 0.380 & -\\
BEVFormer~\cite{bevformer}
    & 0.775 & 0.239 & 0.467\\
\midrule
BEVDepth-Seg
    & 0.827 & \textbf{0.464} & 0.450 \\
MatrixVT
    & \textbf{0.835} & 0.448 & \textbf{0.462}\\
\bottomrule
  \end{tabular}
}
\caption{Experimental results of BEV segmentation on the nuScenes val set. We implement map segmentation on the BEVDepth by placing a segmentation head on the BEV feature.}
\label{table:seg}
	\end{center}
\end{table}

\subsection{Efficient Transformation}
We compare the efficiency of View Transformation in two dimensions: Speed and Memory Consumption. Note that we measure the latency and memory footprint (using fp32) of \emph{view transformers} since these metrics are affected by backbone and head design. We take the CPU as a representative general-purpose device where customized operators are unavailable.
For a fair comparison, we measure and compare the other two Lift-Splat-like view transformers. The LS-BEVDet is the accelerated transformation used in BEVDet (with default parameters); the LS-BEVDepth uses the CUDA~\cite{cuda} operator proposed in BEVDepth~\cite{bevdepth} that is not available on CPU and other platforms. 
To demonstrate the characteristics of different methods, we define six transformation settings, namely S1$\sim$S6, varies in image feature size, BEV feature size, and feature channels that are closely related to model performance.

As can be seen in Fig.~\ref{fig:latency}, the proposed MatrixVT boost the transformation significantly on the CPU, being 4 to 8 times faster than the LS-BEVDet\cite{bevdet}. On the CUDA platform~\cite{cuda}, where customized operators enable faster transformation, MatrixVT still shows a much faster speed than the LS-BEVDepth under most settings. 
Besides, we calculate the number of intermediate variables during VT as an indicator of extra memory footprint. For MatrixVT, these variables include the Ring Matrix, Ray Matrix, and the intermediate matrix; for Lift-Splat, intermediate variables include the intermediate tensor and the predefined BEV grids.
As illustrated in Fig.~\ref{fig:latency} (right), MatrixVT consumes 2 to 15 times less memory than LS-BEVDepth and 40 to 80 times less memory than LS-BEVDet. 

\begin{figure*}
    \centering
    \includegraphics[width=\textwidth]{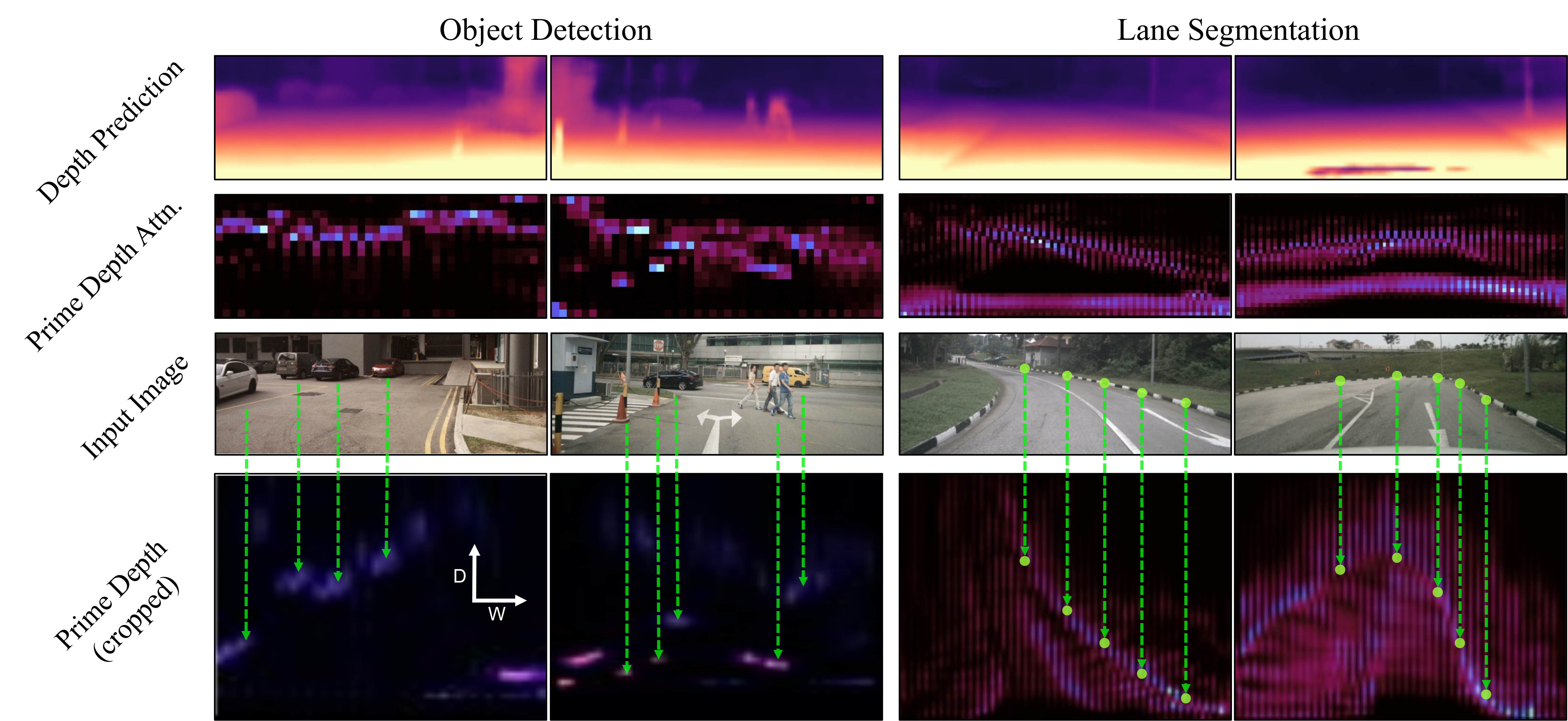}
    \caption{Depth Prediction, Prime Depth Attention, and Prime Depth for object detection and lane segmentation task. The categorical depth prediction is illustrated using distances with largest probability, while the Prime Depth is illustrated by raw probability values.}
    \label{fig:pd}
\end{figure*}

\subsection{Effectiveness of Prime Extraction}
\label{sec:pe-exp}
In this section, we validate the effectiveness of Prime Extraction by performance comparison and visualization.
\subsubsection{Effects on Performance}
As mentioned in Sec.~\ref{sec:compress}, we argue that the features and corresponding depths can be compressed with little or no information loss. 
We thus individually adopt the Prime Extraction onto the BEVDepth~\cite{bevdepth} to verify its effectiveness. Specifically, we compress image features and the depths before the common Lift-Splat. As shown in Tab.~\ref{table:pe}, the BEVDepth with Prime Extraction achieves 41.1\% NDS with Res-50 and 56.1\% NDS with VovNetv2-99~\cite{vovnet2}, which is comparable to the baseline without compression. Thus, we argue that Prime Extraction effectively retained key information from raw features.
\begin{table}
	\begin{center}
		\scalebox{1}{
  \begin{tabular}{c|c|ccc|c}
\toprule
    Backbone & PE & mAP & mATE & NDS & Mem. \\
\midrule
Res-50 & $\times$
    & 0.337 & 0.646 & 0.414 & 734M \\
Res-50 & $\checkmark$
    & 0.336 & 0.644 & 0.411 & \textbf{316M} \\
\midrule
V2-99 & $\times$
    & 0.464 & 0.528 & 0.564 & 2.5G\\
V2-99 & $\checkmark$
    & 0.459 & 0.534 & 0.561 & \textbf{875M}\\
\bottomrule
  \end{tabular}
}
\caption{Effects of using Prime Extraction on BEVDepth. The memory consumption reported is the max memory usage on PyTorch during View Transformation (VT).}
\label{table:pe}
	\end{center}
\end{table}

\subsubsection{Prime Information in Object Detection}
We then delve into the mechanism of Prime Extraction by visualization. Fig.~\ref{fig:pd} shows the inputs and outputs of the Prime Extraction module. It can be seen that the Prime Extraction module trained on different tasks focusing on different information. The Prime Depth Attention in object detection focuses on foreground objects. Thus the Prime Depth retained the depth of objects while ignoring the depth of the background. Also, it can be seen from the yellow car in the second column, which is obscured by three pedestrians. Prime Extraction effectively distinguishes these objects at different depths.

Tab.~\ref{table:pe} shows the effect of adopting Prime Extraction on the BEVDepth~\cite{bevdepth}. We do the per-pixel outer product of the Prime Feature and Prime Depth, then apply Voxel Pooling~\cite{bevdepth} on the obtained tensor. The improved version of BEVDepth saves about 28\% percent of memory consumption while offering comparable performance.

\subsubsection{Prime Information in Map Segmentation}
For the map segmentation task, we take lane segmentation as an example --- the Prime Extraction module focus on lane and road that is closely related to this task. However, since the area of the target category is a wide range covering several depth bins, the distribution of Prime Depth is uniform in the target area (see Fig.~\ref{fig:pd}, 3rd and 4th columns). 
The observation indicates that Prime Extraction generates a new form of depth distribution that fits the map segmentation task. With the Prime Depth that is rather uniform, the same feature can be projected to multiple depth bins since they are occupied by the same category.

\subsection{Extraction of Prime Feature}
In the PFE, we propose using max pooling followed by several 1D convolutions to reduce and refine the image feature. Before reduction, the coordinate of each pixel is embedded into the feature as position embedding. We conduct experiments to validate the contribution of each design.

A possible alternative of the max pooling is the CollapseConv as in \cite{caddn} and \cite{pon}, which merges the height dimension into the channel dimension and reduces the merged channel by linear projection. However, the design of CollapseConv brings some disadvantages. For example, the merged dimension is of size $C \times H$, which is high and requires extra memory transportation. To address these problems, we propose using max pooling to reduce the image feature in Prime Extraction. Tab.~\ref{table:reduce} shows that reduction using max pooling achieves even better performance than CollapseConv while eliminating these shortcomings.

We also conduct experiments to show the effectiveness of the refine sub-net after reducing in Tab.~\ref{table:reduce}. The results indicate that the refine sub-net plays a vital role in adapting the reduced feature to BEV space, without which a performance drop of 0.9\% mAP will occur.
Finally, the experimental results in Tab.~\ref{table:reduce} have shown that the position embedding brings an improvement of 0.4\% mAP, which is also preferable.
\begin{table}
	\begin{center}
		\scalebox{1}{
  \begin{tabular}{c|cc|cc}
\toprule
    Reduction & Refine & Pos. Emb. & mAP & NDS \\
\midrule
Max Pooling & $\times$ & $\times$
    & 0.328 & 0.393 \\
Max Pooling & $\checkmark$ & $\times$
    & 0.332 & 0.414 \\
Max Pooling & $\checkmark$ & $\checkmark$
    & 0.336  & 0.415 \\
CollapseConv & $\checkmark$ & $\checkmark$
    & 0.334 & 0.404 \\
\bottomrule
  \end{tabular}
}
\caption{Performance of MatrixVT with different feature reduction methods. The ``Refine'' indicates the refine convs, the ``Pos.Emb.'' indicates using front-view position embedding.}
\label{table:reduce}
	\end{center}
\end{table}

\section{Conclusion}
This paper proposes a new paradigm of View Transformation (VT) from multi-camera to Bird's-Eye-View. The proposed method, MatrixVT, generalizes VT into a feature transportation matrix. We then propose Prime Extraction, which eliminates the redundancy during transformation, and the Ring \& Ray Decomposition, which simplifies and boost the transformation. While being faster and more efficient on both specialized devices like GPU and general-purpose devices like CPU, our extensive experiments on the nuScenes benchmark indicate that the MatrixVT offers comparable performance to the state-of-the-art method.
\newpage
\newpage
{\small
\bibliographystyle{ieee_fullname}
\bibliography{matrixvt}
}
\clearpage
\section*{Appendix}
\subsection*{A1.0 Symbol Definitions}
Here, we give a brief description of symbols used in the below sections in Tab.~\ref{tab:symbols}.
\begin{table}[h]
    \centering
    \scalebox{0.9}{
    \begin{tabular}{c|c|c}
    \toprule
    Symbol & Definition & Example \\
    \midrule
    $N_c$ & number of cameras for a scene & 6 \\
    $N_d$ & number of depth bins & 112 \\
    $W_I$ & width of image features & 44 \\
    $H_I$ & height of image features & 16 \\
    $W_B$ & width of BEV features (\#grids) & 128 \\
    $H_B$ & height of BEV features (\#grids) & 128 \\
    $C$   & number of feature channels & 80\\
    \midrule
    $S$   & $H_B \times W_B$ & $128 \times 128$\\
    $W$   & $N_c \times W_I$ & $6 \times 44$\\
    \midrule
    $\mathrm{F}_I$ & Prime Features & - \\
    $\mathrm{D}$ & Prime Depths & - \\
    $\mathrm{F}_{BEV}$ & BEV features & - \\
    
    \bottomrule
    \end{tabular}
    }
    \caption{Definition of symbols.}
    \label{tab:symbols}
\end{table}

Besides, letters of the normal script (\eg $F$) denote tensors, letters of Roman script (\eg $\mathrm{M}$) denote matrices, $\mathrm{A} \cdot \mathrm{B}$ denotes matrix multiplication, $\mathrm{A} \odot \mathrm{B}$ denotes Hadamard Product.

\subsection*{A1.1 Generation of Ring \& Ray Matrices}
\begin{algorithm}
\caption{Generation of Ring \& Ray Matrices}
\label{alg:gen}
\hspace*{0.02in} {\bf Input:}\\
\hspace*{0.1in}{Image Feature Geometry: $G_I \in \mathcal{R}^{N_c \times W_I \times N_d \times 2}$;\\}
\hspace*{0.1in}{BEV Grids: $G_B \in \mathcal{R}^{H_B \times W_B \times 4}$\\}
\hspace*{0.02in} {\bf Output:}\\
\hspace*{0.1in}{Ring Matrix: $\mathrm{M_{Ring}} \in \mathcal{R}^{(H_B \times W_B) \times N_d}$\\}
\hspace*{0.1in}{Ray Matrix: $\mathrm{M_{Ray}} \in \mathcal{R}^{(H_B \times W_B) \times (N_c \times W_I)}$}\\
{\noindent}\rule[5pt]{\columnwidth}{0.05em}
\begin{algorithmic}
\State $M_{Ring} = \mathbb{0}^{H_B \times W_B \times N_d}$,
\State $M_{Ray} = \mathbb{0}^{H_B \times W_B \times N_c \times W_I}$.
    \For{($h_b= 1 \rightarrow H_B$, $w_b= 1 \rightarrow W_B$)}
        \For{($n = 1 \rightarrow N_c$, $w_i = 1 \rightarrow W_I$, $d = 1 \rightarrow N_d$)}
        \If{$G_i(n,w_i,d)$ in $G_b(h_b,w_b)$}
            \State{$M_{Ring}(h_b,w_b,d)=1$}
            \State{$M_{Ray}(h_b,w_b,n,w_i)=1$}
            \EndIf
        \EndFor
    \EndFor
\State $\mathrm{M_{Ring}}$ = reshape($M_{Ring}$,$(H_B \times W_B) \times N_d$)
\State $\mathrm{M_{Ray}}$ = reshape($M_{Ring}$,$(H_B \times W_B) \times (N_c \times W_I)$)
\State \Return $\mathrm{M_{Ring}},\mathrm{M_{Ray}}$
\end{algorithmic}
\end{algorithm}
The generation of the Ring Matrix and the Ray Matrix relies on the intrinsic and extrinsic parameters of the camera setting. These parameters determine the geometrical relationship between the ``lifted'' features and the BEV grids. Note that \emph{these matrices need to be generated only once for real-world applications} - as the camera positions are usually fixed. 

To simplify the algorithm, we take the geometry (2D coordinate, $(x,y)$) of the ``lifted'' Prime Feature and the BEV grids (2D grids, $(x_0,y_0,x_1,y_1)$), instead of intrinsic and extrinsic parameters as input. 
One of the algorithms that generate these matrices is described in Alg.~\ref{alg:gen}.

\subsection*{A1.2 Pipeline Reformulation}
In MatrixVT, we reformulate our pipeline into a new form to eliminate huge intermediate tensors. Now we prove that these two forms are mathematically equivalent. For clarity, we mark the shape of variables in their upper right corner.

We omit the $C$ in the below equations for simplicity, the Prime Feature $\mathrm{F}_{BEV}^{S \times C}$ is therefore viewed as a vector $\boldsymbol{f}_{BEV}^{S \times 1}$. 
Since the ``lift'' operation is a per-pixel outer product of the feature and categorical depth, we rewrite the whole pipeline as follows:
\begin{align}
    &\boldsymbol{f}_{BEV}^{S \times 1} = (\mathrm{M}_{Ray}^{S \times W} \odot \mathrm{F}_{inter}^{S \times W}) \cdot \mathbb{1}^{W \times 1}\label{eq:1}\\
    &\mathrm{F}_{inter}^{S \times W} =
    \mathrm{M}_{Ring}^{S \times N_d}
    \cdot \mathrm{D}^{N_d \times W} \cdot \mathrm{F}_I^{W \times W},
    \label{eq:2}
\end{align}
where $\mathrm{F}_I^{W \times W}$ denotes a diagonal matrix that satisfies $diag(\mathrm{F}_I^{W \times W}) = \boldsymbol{f}_I^{W \times 1}$.

Taking Eq.~\ref{eq:2} into Eq.~\ref{eq:1}, the prime feature $\boldsymbol{f}_{BEV}$ can be derived as: 
\begin{equation}
\begin{aligned}
    \boldsymbol{f}_{BEV}^{S \times 1} &= (\mathrm{M}_{Ray}^{S \times W} \odot \mathrm{F}_{inter}^{S \times W}) \cdot \mathbb{1}^{W \times 1}\\
    &=\mathrm{M}_{Ray}^{S \times W} \odot \left(\mathrm{M}_{Ring}^{S \times N_d} \cdot \mathrm{D}^{N_d \times W} \cdot \mathrm{F}_I^{W \times W}\right) \cdot \mathbb{1}^{W \times 1}\\
    &=  \left(\mathrm{M}_{Ray}^{S \times W} \odot \left(\mathrm{M}_{Ring}^{S \times N_d} \cdot \mathrm{D}^{N_d \times W}\right)\right) \cdot \boldsymbol{f}_I^{W \times 1}\\
    &= \mathrm{M}_{FT}^{S \times W} \cdot \boldsymbol{f}^{W \times 1}_I,
\end{aligned}
\end{equation}
where $\mathrm{M}_{FT}^{S \times W}=\left(\mathrm{M}_{Ray}^{S \times W} \odot \left(\mathrm{M}_{Ring}^{S \times N_d} \cdot \mathrm{D}^{N_d \times W}\right)\right)$.

Since the size of $\mathrm{M_{FT}}$ is invariant to the feature channels, the memory footprint during transformation is saved.
\end{document}